\documentclass[10pt,twocolumn,letterpaper]{article}

\usepackage[pagenumbers]{cvpr} 

\usepackage{graphicx}
\usepackage{amsmath}
\usepackage{amssymb}
\usepackage{booktabs}
\usepackage{algorithm}
\usepackage{lipsum}
\usepackage{url}

\usepackage{mathtools}

\usepackage{tabularx}
\usepackage{multirow}

\usepackage{trimclip}

\usepackage{listings}
\usepackage{upquote}  
\usepackage{xcolor,colortbl}
\lstdefinestyle{overleaf}{
    backgroundcolor=\color[rgb]{0.95,0.95,0.92},   
    commentstyle=\color[rgb]{0,0.6,0},
    keywordstyle=\color{magenta},
    numberstyle=\tiny\color[rgb]{0.5,0.5,0.5},
    stringstyle=\color[rgb]{0.58,0,0.82},
    basicstyle=\ttfamily\footnotesize,
    breakatwhitespace=false,         
    breaklines=true,                 
    captionpos=b,                    
    keepspaces=true,                 
    numbers=left,                    
    numbersep=5pt,                  
    showspaces=false,                
    showstringspaces=false,
    showtabs=false,                  
    tabsize=2
}
\lstdefinestyle{mocov3}{
  backgroundcolor=\color{white},
  basicstyle=\fontsize{7.5pt}{7.5pt}\ttfamily\selectfont,
  columns=fullflexible,
  breaklines=true,
  captionpos=b,
  commentstyle=\fontsize{7.5pt}{7.5pt}\color[rgb]{0.25,0.5,0.5},
  keywordstyle=\fontsize{7.5pt}{7.5pt}\color[rgb]{0.85,0.18,0.50},
}
\usepackage{etoolbox}
\makeatletter
\AfterEndEnvironment{algorithm}{\let\@algcomment\relax}
\AtEndEnvironment{algorithm}{\kern2pt\hrule\relax\vskip3pt\@algcomment}
\let\@algcomment\relax
\newcommand\algcomment[1]{\def\@algcomment{\footnotesize#1}}
\renewcommand\fs@ruled{\def\@fs@cfont{\bfseries}\let\@fs@capt\floatc@ruled
  \def\@fs@pre{\hrule height.8pt depth0pt \kern2pt}%
  \def\@fs@post{}%
  \def\@fs@mid{\kern2pt\hrule\kern2pt}%
  \let\@fs@iftopcapt\iftrue}
\makeatother

\usepackage[pagebackref,breaklinks,colorlinks]{hyperref}

\usepackage[capitalize]{cleveref}
\crefname{section}{Sec.}{Secs.}
\Crefname{section}{Section}{Sections}
\Crefname{table}{Table}{Tables}
\crefname{table}{Tab.}{Tabs.}

\def\cvprPaperID{0000} 
\def\confName{CVPR}
\def\confYear{2023}

\newcommand{\authsep}{\;\;}

\usepackage{bbding}   
\usepackage{xcolor}   
\definecolor{nicegreen}{HTML}{009900} 
\definecolor{nicered}{HTML}{CC0000}   
\newcommand{\greencheck}{\textcolor{nicegreen}{\Checkmark}}
\newcommand{\redcross}{\textcolor{nicered}{\XSolidBrush}}

\usepackage{colortbl} 

\usepackage{ragged2e}

\usepackage{amssymb}
\usepackage{url}
\usepackage{hyperref}

\begin{document}


\title{Tooth Structure-aware Prior and Lesion-aware Dynamic Loss Refinement for DETR Based Caries Detection}

\author{Xuefen Liu$^{1}$ \authsep Xinquan Yang$^{2}$ \authsep Mianjie Zheng$^1$ \authsep Kun Tang$^1$ \\
Xuguang Li$^3$ \authsep Xiaoqi Guo$^1$ \authsep Linlin Shen$^{2\star}$ \authsep He Meng$^3$ }
\maketitle
{\let\thefootnote\relax\footnote{{$^{\star}$ Corresponding author. \\
$^{\quad\;\;\ 1}$ College of Computer Science and Software Engineering, Shenzhen University, Shenzhen, China. \\
$^{\quad\;\;\ 2}$ School of Artificial Intelligence, Shenzhen University, Shenzhen, China.\\
$^{\quad\;\;\ 3}$ Department of Stomatology, Shenzhen University General Hospital, Shenzhen, China.}}}


\vspace{-1em}

\maketitle

\begin{abstract}
	As dental caries appear as subtle, low-contrast lesions in intraoral imaging, existing deep learning models face significant challenges in the early detection of caries. While recent Transformer-based detectors have shown promising results in natural images, they often fail to capture the domain-specific anatomical priors crucial for dental caries detection. In this paper, we propose Caries-DETR, a specialized Transformer framework for caries detection in intraoral images. A Tooth Structure-aware Query Initialization (TSQI) is designed, leveraging large-scale intraoral photograph pre-training and a structure perception branch (SPB) to integrate high-frequency structural priors, guiding the model to focus on anatomically significant lesion areas. Furthermore, we design a Lesion-aware Dynamic Loss Refinement (LDLR) to implement quality-driven hard mining through adaptive loss reweighting based on lesion size, anatomical relevance, and prediction quality, optimizing detection for subtle lesions. Extensive experiments on two public datasets (i.e., AlphaDent and DentalAI) demonstrate that Caries-DETR achieves a state-of-the-art performance compared to existing methods and exhibits good generalization and robustness. Code and data at \href{https://github.com/XuefenLiu-SZU/Caries-DETR}{https://github.com/XuefenLiu-SZU/Caries-DETR}.
\end{abstract}


\section{Introduction}
Dental caries, the localized destruction of dental hard tissues caused by acidic by-products from bacterial fermentation of dietary carbohydrates, remains a major global public health problem affecting billions of people worldwide~\cite{kassebaum2015global}. 
Early detection is crucial as it facilitates remineralization therapies and minimally invasive interventions, thereby preserving tooth structure and reducing the need for extensive restorative care~\cite{fejerskov2015dental}. 
In recent years, intraoral clinical photography is favored due to its low cost, non-invasiveness, and ease of acquisition~\cite{desai2013digital,teruya2025dental}.
These images provide direct visualization of the surfaces of the teeth, which enable the association of subtle changes in color and texture with early demineralization~\cite{kaczmarek2005digital}. 
However, manual interpretation of these images is time-consuming and subject to inter- and intra-observer variability.
Deep learning-based computer-aided detection (CAD) systems have shown great potential in the automatic detection of dental caries in radiographs and near-infrared transillumination images due to their efficiency and high accuracy~\cite{lee2018detection}.
Despite these advancements, developing robust and accurate deep learning models for caries detection on intraoral images remains challenging. 
This challenge stems primarily from the inherent variability due to differences in lighting, camera angles, and patient anatomy~\cite{ku2025accuracy}, and the subtle, often indistinct, visual appearance of early carious lesions~\cite{gomez2015detection}.

To address these challenges, researchers have developed specialized networks for dental caries detection, mainly with CNN-based and Transformer-based paradigms. 
Most CNN-based methods~\cite{ren2015faster,lin2017focal} rely on predefined anchor boxes, which require heuristic design and often fail to adapt to the highly variable morphology of carious lesions, leading to suboptimal detection of subtle lesions and irregular boundaries. 
In contrast, Transformer-based methods~\cite{carion2020end,zhu2021deformable} eliminate the need for handcrafted anchors and non-maximum suppression by formulating detection as an end-to-end set prediction problem. 
However, these methods typically suffer from slow convergence and poor performance on small caries lesions. 
A key reason is that their query initialization strategies, often random or derived from natural image statistics, overlook a critical characteristic of intraoral imagery: structural clues. 
These structural patterns are actually more stable under varying illuminations. The failure to incorporate these domain-specific structural priors during query initialization limits their effectiveness in localizing small, subtle lesions in dental images~\cite{zhang2022dino}.


To address this limitation, we propose a \textbf{Tooth Structure-aware Query Initialization (TSQI)}, which provides the dental prior to facilitate the network convergence. 
TSQI was first pre-trained in a self-supervised manner on a large-scale intraoral dataset, using high-frequency gradient maps as supervisory signals. 
During finetuning, it generates a structural saliency map that highlights structurally salient and lesion-prone regions. 
Then, this map is used to modulate the selection of feature locations for query initialization, thereby anchoring object queries to anatomically meaningful areas and prioritizing the localization of subtle structural anomalies. 
By integrating these learned structural priors into the transformer’s query generation process, TSQI effectively guides the model’s attention toward anatomically significant and lesion-prone regions.

Another key limitation of existing transformer-based methods lies in the severe imbalance between positive and negative samples during training. Early-stage caries are often small and sparsely distributed, leading to a skewed optimization process where challenging positive samples are easily neglected. Although prior works have attempted to address this imbalance~\cite{meng2021conditional,liu2022dab,huang2024dq}, they perform poorly on caries detection because they fail to account for the specific visual characteristics of caries lesions—such as low density, irregular boundaries, and subtle enamel changes—which differ substantially from those of generic small objects.
To overcome this issue, we design a \textbf{Lesion-aware Dynamic Loss Refinement (LDLR)}, which implements a lesion-aware supervision strategy. LDLR adaptively re-weights the regression loss based on lesion size, anatomical relevance, and prediction quality. Through this dynamic re-weighting, small, low-contrast, or poorly localized caries are assigned higher penalty weights during training, which stabilizes the optimization process and accelerates the model’s convergence toward clinically accurate lesion boundaries.

In summary, the key contributions of this paper are as follows:

\begin{itemize} 
	\item We design a Tooth Structure-aware Query Initialization module, which was pretrained on a large-scale intraoral dataset and integrates the structural priors into the transformer's query generation process, guiding the model's attention towards anatomically significant and lesion-prone regions. 
	\item We develop the Lesion-aware Dynamic Loss Refinement, an adaptive optimization module that performs quality-driven hard mining by modulating gradients based on lesion size, anatomical relevance, and prediction fidelity, significantly improving performance on challenging lesions. 
	\item Extensive experiments on two public intraoral datasets (AlphaDent and DentalAI), suggest that our Caries-DETR achieves superior performance and clinical relevance compared to state-of-the-art methods. 
\end{itemize}


\section{Related work}

\subsection{Object Detection}
Object detection is a fundamental computer vision task, and numerous architectures have been proposed~\cite{carion2020end}. Ren et al.~\cite{ren2015faster} introduced Faster R-CNN, a pioneering two-stage detector that integrates a region proposal network with a fast R-CNN detector. 
Lin et al.~\cite{lin2017focal} proposed RetinaNet, a single-stage detector that addresses the class imbalance problem using a focal loss. Zhou et al.~\cite{zhou2019objects} introduced CenterNet, a method that models objects as center points, simplifying the detection process.
YOLO~\cite{farhadi2018yolov3,tian2025yolov12} formulates object detection as a unified regression problem, enabling end-to-end optimization and fast inference. 

The Transformer architecture~\cite{vaswani2017attention} has fundamentally reshaped object detection by enabling global modeling through attention mechanisms. 
Carion et al.~\cite{carion2020end} proposed DETR, an end-to-end object detection framework with Transformers that eliminates the need for handcrafted components like non-maximum suppression and anchor generation. 
While innovative, DETR suffers from slow convergence and poor performance on small objects~\cite{carion2020end,zhang2022dino}. 
Zhu et al.~\cite{zhu2021deformable} addressed these issues with Deformable DETR, which introduces deformable attention modules for faster convergence and better performance. 
Zhang et al.~\cite{zhang2022dino} proposed DINO, a state-of-the-art DETR-based model that improves denoising anchor boxes for end-to-end object detection. 
Zong et al.~\cite{zong2023detrs} proposed Co-DETR, which coordinates multiple heads for better object detection. 
Huang et al.~\cite{huang2025deimv2} introduced DEIMv2, which improves matching efficiency and training stability through refined instance modeling strategies. Zhang et al.~\cite{zhang2024mr} proposed Mr.DETR, emphasizing multi-resolution feature alignment to enhance detection performance across object scales. Robinson et al.~\cite{robinson2025rfdetrneuralarchitecturesearch} presented RF-DETR, which leverages neural architecture search to automatically design efficient DETR-style architectures, achieving favorable accuracy–efficiency trade-offs.


\subsection{Deep Learning in Caries Detection}
Deep learning has increasingly been applied to dental caries detection from intraoral photographs, which offer non-invasive acquisition and suitability for tele-dentistry but exhibit substantial variability in illumination, occlusion, and background clutter. 
Early studies predominantly relied on convolutional neural networks (CNNs) to analyze oral images. 
Building on these efforts, Saini et al.~\cite{saini2021dental} evaluated multiple CNN backbones, including ResNet-50, for early caries classification in tele-dentistry scenarios. 
Ding et al.~\cite{ding2021detection} further extended CNN-based approaches to lesion localization by adopting YOLOv3 for caries detection in mobile oral images. 
Jiang et al.~\cite{jiang2023cariesfg} proposed CariesFG, a fine-grained classification framework that integrates a coordinate attention mechanism to suppress background noise and highlight lesion-specific features. 

Improving model transparency and robustness against data heterogeneity has been another focal point. 
Kim et al.~\cite{kim2025cnn} introduced ResFC, a remote diagnostic model utilizing a Region-Based Autoencoder (RB-AE) and Gradient-weighted Class Activation Mapping (Grad-CAM) to provide visual explanations for model predictions. 
Similarly, Asghar et al.~\cite{asghar2025cariesxplainer} presented CariesXplainer, which employs a lightweight MobileNetV3 backbone to enhance interpretability while maintaining computational efficiency. 
To further address the issue of label ambiguity in dental datasets, Zhang et al.~\cite{zhang2025multi} devised M3C, a multi-category fusion contrastive learning framework. 
By implementing a core data selection strategy, they significantly improved feature robustness against noisy labels in RGB-based dental classification.
Recognizing the limitations of local convolution operators, recent studies have begun exploring Transformers for global feature modeling. 
Asif et al.~\cite{asif2025oraltransnet} proposed OralTransNet, a hybrid architecture that fuses CNN-extracted local features with Transformer-based global attention mechanisms to improve diagnostic accuracy for various oral diseases. 
More recently, Liu et al.~\cite{liu2026enhancing} incorporated Agent Attention into Vision Transformers (ViT) within a Conformer architecture, aiming to further refine the interaction between global and local contexts.
While the aforementioned works have advanced the field, they exhibit specific limitations when applied to high-precision caries detection. 
Hybrid models like OralTransNet~\cite{asif2025oraltransnet} often rely on lightweight backbones to balance computation, which inherently restricts feature capacity for high-precision detection tasks.


\section{Method}
\begin{figure*}[!h]
	\centering
	\includegraphics[width=1\linewidth]{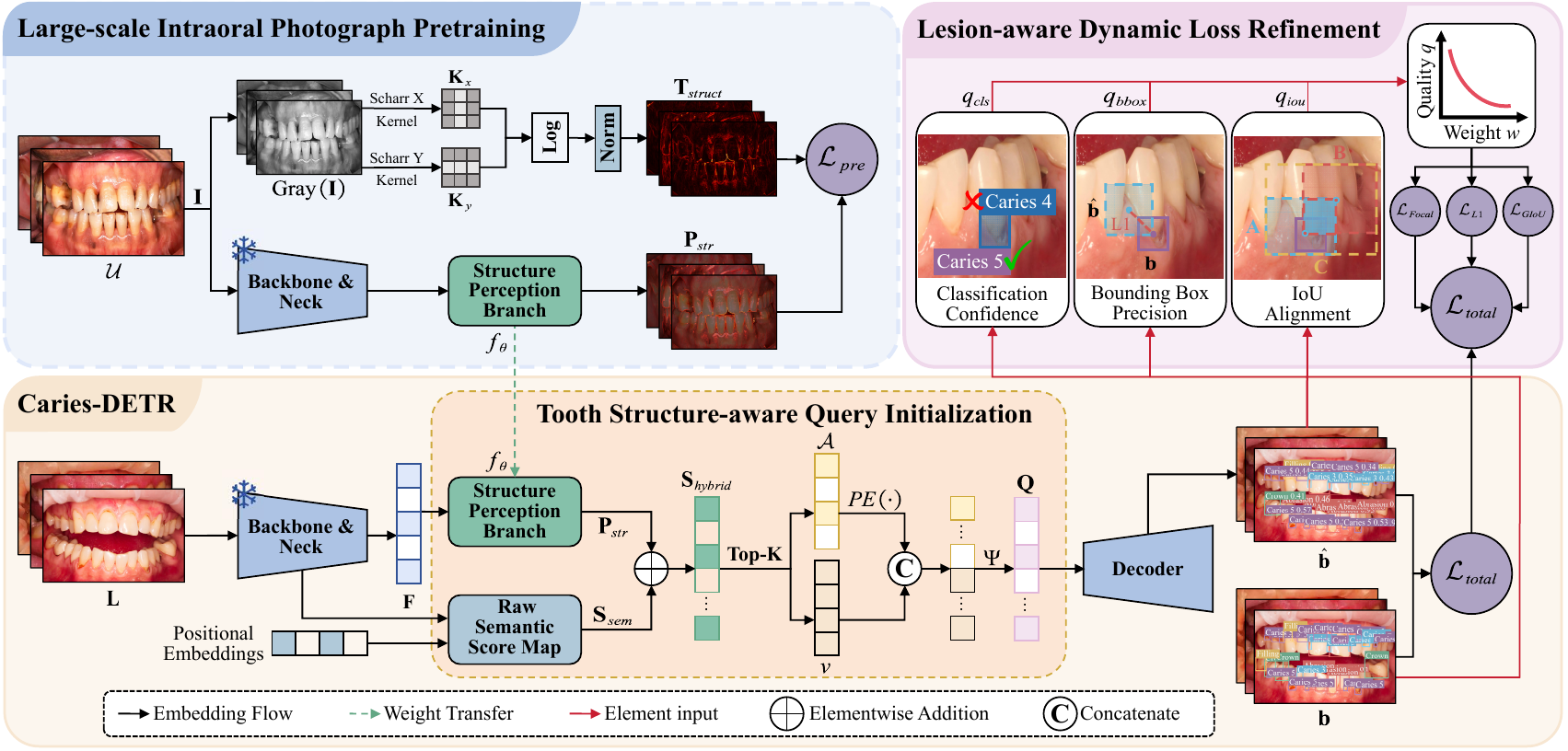}
	\caption{Overall framework of the proposed Caries-DETR. 
	}
	\label{fig:framework}
\end{figure*}

\subsection{Tooth Structure-aware Query Initialization}
In Detection Transformers (DETR), object queries act as learnable ``searchlights'' that interact with image features via cross-attention to localize and classify instances. Standard DETR frameworks typically use fixed learnable embeddings or simple feature-sampling for query initialization. However, these methods are purely appearance-driven and ignore a fundamental property of intraoral scenes: carious lesions are inherently coupled with dental anatomy. Pathological indicators—such as lesions in pits, fissures, and interproximal contact areas—depend more on tooth-specific structural patterns than on inconsistent photometric cues like saliva reflections or specular highlights.

To address this limitation, we propose a Tooth Structure-aware Query Initialization (TSQI) method. Unlike generic initialization, TSQI explicitly encodes tooth structural priors to guide the model’s attention to clinically relevant regions. This domain-specific guidance alleviates the burden on the model to learn complex oral anatomy from scratch, thereby enhancing learning efficiency and detection precision.
The proposed TSQI consists of two main steps:\par

\textbf{1) Large-scale Intraoral Photograph Pretraining.} 
We first collect a large dataset of intraoral photographs, denoted as $\mathcal{U}$, which comprises an unlabeled corpus of 30,943 diverse intraoral images aggregated from various public sources, 
as summarized in the supplementary material (Table~S1).
This dataset covers patients of different ages and is collected from multiple centers to ensure diversity. 
We then devise a gradient-guided self-supervised pretext task to explicitly encode high-frequency geometric primitives that are crucial for dental diagnosis.
Specifically, we employ the Scharr operator for gradient computation, which is preferred over simpler approximations due to its superior rotational symmetry and sensitivity to fine-grained textural changes. 
For each raw intraoral image $\mathbf{I} \in \mathbb{R}^{C\times H\times W}$, we first convert it to grayscale and compute its gradient magnitude map:
\begin{equation}
	\mathbf{G}(\mathbf{I}) = \sqrt{(\text{Gray}(\mathbf{I})  \mathbf{K}_x)^2 + (\text{Gray}(\mathbf{I})  \mathbf{K}_y)^2},
\end{equation}
where $\mathbf{K}_x$ and $\mathbf{K}_y$ are the Scharr kernels.
To address challenges such as variable illumination and specular reflections inherent in intraoral photography, we apply logarithmic compression followed by instance-level normalization to generate a robust pseudo-ground-truth structural target:
\begin{equation}
	\mathbf{T}_{struct} = \text{Norm}(\log(1 + \mathbf{G})).
\end{equation}
This transformation amplifies low-contrast structural cues while suppressing high-intensity noise.
After extracting $\mathbf{T}_{struct}$ from $\mathbf{I}$, we introduce a Structure Perception Branch (SPB), denoted as $f_\theta$, which is a lightweight convolutional neural network designed to map features from the backbone's neck to the structural space. The SPB is trained on $\mathcal{U}$ by minimizing the pixel-wise $L_1$ reconstruction loss:
\begin{equation}
	\mathcal{L}_{pre} = \frac{1}{|\mathcal{U}|} \sum_{\mathbf{I} \in \mathcal{U}} \| f_\theta(\text{Backbone}(\mathbf{I})) - \mathbf{T}_{struct}(\mathbf{I}) \|_1.
\end{equation}
This pre-training ensures that $f_\theta$ learns to robustly perceive and localize critical dental structures, thereby establishing a powerful domain-specific prior for the subsequent supervised detection task. 

\textbf{2) Tooth Structure-aware Query Initialization.} 
The pre-trained $f_\theta$ is directly integrated into the training of Caries-DETR.
During training, SPB generates a structural saliency map $\mathbf{P}_{str}$ from $\mathbf{F}$:
\begin{equation}
	\mathbf{P}_{str} = \sigma(f_\theta(\mathbf{F})),
\end{equation}
where $\sigma$ is the sigmoid activation. 
$\mathbf{P}_{str}$ highlights structurally significant and lesion-prone areas.
We formulate a hybrid attention score $\mathbf{S}_{hybrid}$ by performing element-wise modulation of semantic and structural cues:
\begin{equation}
	\mathbf{S}_{hybrid} = \mathbf{S}_{sem} \odot (\mathbf{1} + \lambda \cdot \mathbf{P}_{str}),
\end{equation}
where $\lambda$ is a learnable scaling factor, and $\mathbf{S}_{sem}$ is a raw semantic score map extracted from the backbone.
Through this Hadamard product, the scores of anchor points that possess both semantic relevance and strong structural evidence are amplified. This effectively guides the model's initial focus towards regions that are clinically significant yet visually ambiguous in raw pixel space.
For the top-$K$ anchor points $\mathcal{A} = \{(x_i, y_i)\}_{i=1}^K$ selected from $\mathbf{S}_{hybrid}$, we further enrich their representations by constructing confidence-embedded positional queries. 
Specifically, for each anchor point $(x_i, y_i)$, we define a query embedding $\mathbf{Q}_i$ that explicitly integrates its spatial location and a lesion-aware confidence measure.
Let $v_i = \mathbf{S}_{hybrid}^{(x_i, y_i)}$ denote the hybrid attention score at $(x_i, y_i)$, which reflects the joint semantic and structural significance of the anchor point. We first apply a positional encoding function $PE(\cdot)$ to map the normalized coordinates $(x_i, y_i)$ into a positional embedding. This embedding is then concatenated with the corresponding score $v_i$ and projected into a high-dimensional query space:
\begin{equation}
	\mathbf{Q}_i = \Psi\left( \text{Concat}\left( PE(x_i, y_i), \ v_i \right) \right),
\end{equation}
where $\Psi$ is a linear projection layer that aligns the combined representation with the transformer's query dimension.

By encoding $v_i$ directly into the query, the transformer decoder receives richer contextual information. This design enables the attention mechanism to prioritize high-confidence proposals and effectively suppress false positives arising from background noise or visually similar but diagnostically irrelevant textures.

\subsection{Lesion-aware Dynamic Loss Refinement}
Despite advancements in feature representation and query initialization, caries detection in intraoral photographs remains challenging due to the extreme class imbalance between positive (caries) and negative (sound tissue) samples. 
This challenge is compounded by the prevalence of ``hard'' instances—such as small, occluded, or low-contrast lesions—across different caries types. 
During training, gradients are often dominated by easy-to-classify samples, causing these hard instances to be inadequately learned. 
To address this issue, we propose a lesion-aware dynamic loss refinement (LDLR), which implements a quality-driven hard mining strategy. The design of LDLR is meticulously guided by the specific clinical requirements of dental diagnosis, where accurate prediction demands a comprehensive evaluation across spatial, semantic, and geometric dimensions.
Specifically, we define a quality vector $\mathbf{q} = [q_{cls}, q_{bbox}, q_{iou}]^T$, which assesses the prediction reliability from three orthogonal perspectives tailored to the challenges of dental imaging.

\noindent \textbf{1) Classification Confidence ($q_{cls}$):} The visual distinction between different stages of caries is often subtle, leading to significant semantic ambiguity. A low classification probability indicates that the model struggles to capture the discriminative features necessary for accurate pathological grading. We therefore utilize the classification score $q_{cls}$ to measure the model's confidence in category identification, which encourages the network to focus on hard-to-distinguish lesion types.\\ 
\noindent \textbf{2) Bounding Box Precision ($q_{bbox}$):} Unlike general objects, dental lesions require high-precision coordinate regression for treatment planning. Standard overlap metrics can be insensitive to minor boundary shifts. To address this, we introduce $q_{bbox}$ to quantify localization fidelity based on the L1 distance between the predicted and ground-truth box coordinates. This metric enforces strict geometric alignment by penalizing predictions that—even with decent overlap—fail to precisely delineate the lesion's irregular contours.\\ 
\noindent \textbf{3) IoU Alignment ($q_{iou}$):} In the dense oral environment where teeth are tightly packed, even a slight spatial deviation can lead to the erroneous attribution of a lesion to an adjacent healthy tooth. We employ the Intersection-over-Union $q_{iou}$ as a stringent metric to assess structural overlap. This ensures that the predicted region not only has correct coordinates but also maintains substantial spatial intersection with the ground truth, thereby minimizing false positives on neighboring dental structures.

\paragraph{Dynamic Weight Formulation and Optimization.}
To translate these metrics into dynamic loss weights, we define a linear hardness penalty function $\omega: [0, 1] \to [1, 1+\eta]$:
\begin{equation}
	\omega(q; \eta) = 1 + \eta \cdot (1 - q),
\end{equation}
where $\eta$ is a sensitivity coefficient controlling the emphasis on hard samples. 
As the prediction quality $q$ decreases, the corresponding weight increases linearly, amplifying the gradient contribution during optimization. 
Accordingly, the dynamic weights for each loss component are defined as:
$ w_{cls}^{(i)} = \omega(q_{cls}^{(i)}; \alpha)$, $w_{bbox}^{(i)} = \omega(q_{bbox}^{(i)}; \beta)$, $w_{iou}^{(i)} = \omega(q_{iou}^{(i)}; \gamma)$,
where $\alpha$, $\beta$, and $\gamma$ are task-specific sensitivity hyperparameters. 
The final training objective $\mathcal{L}_{total}$ is a dynamically weighted summation over the set of positive matches $\Omega_{pos}$:
\begin{equation}
	\begin{split}
		\mathcal{L}_{total} = 
		\sum_{i \in \Omega_{pos}} \Big[
		&w_{cls}^{(i)} \mathcal{L}_{Focal}(p_{i,y_i}, y_i)\\
		&+ w_{bbox}^{(i)} \mathcal{L}_{L1}(\hat{\mathbf{b}}_i, \mathbf{b}_i) \\
		&+ w_{iou}^{(i)} \mathcal{L}_{GIoU}(\hat{\mathbf{b}}_i, \mathbf{b}_i)
		\Big],
	\end{split}
\end{equation}
where $p_{i,y_i}$ represents the predicted probability for the ground truth class $y_i$, and $\hat{\mathbf{b}}_i$ and $\mathbf{b}_i$ denote the predicted and ground truth bounding box coordinates, respectively.

This adaptive weighting mechanism naturally induces an automatic curriculum learning process: in early training stages, prevalent low-quality predictions result in high overall weights, accelerating initial convergence. As the model matures and prediction quality improves, the weights gradually decay, allowing for stable, fine-grained refinement of detection boundaries.

\subsection{Caries-DETR}
Dental caries lesion area detection is fundamentally an object detection task. Conventional detectors often underperform on this task due to the substantial domain shift between natural images and intraoral images, as well as the intricate texture characteristics of early caries lesions. To overcome these limitations, we propose Caries-DETR, an end-to-end transformer-based framework for accurate caries detection. An overview of the proposed Caries-DETR architecture is shown in Fig.~\ref{fig:framework}.

The Caries-DETR framework processes a labeled intraoral image $\mathbf{L} \in \mathbb{R}^{C\times H\times W}$ through a backbone network to extract a set of multi-scale feature maps $\mathbf{F}$. 
These features are then fed into a pretrained structure perception module $f_\theta(\mathbf{\cdot})$, which generates a robust structural prior $\mathbf{P}_{str}$ representing dental anatomy. 
Subsequently, semantic features $\mathbf{S}_{sem}$ are produced by incorporating positional embeddings into $\mathbf{F}$. 
A hybrid attention map $\mathbf{S}_{hybrid}$ is formed by performing element-wise modulation between $\mathbf{S}_{sem}$ and $\mathbf{P}_{str}$.
From $\mathbf{S}_{hybrid}$ , the top-K anchor points $\mathcal{A}$ and their corresponding hybrid activation scores $v_i$ are selected and fused using a learnable linear mapping $\Psi$. 
This results in a query embedding $\mathbf{Q}_i$ that integrates both spatial location details and lesion-aware confidence information. 
The query embeddings are then passed to the decoder for final caries lesion detection. During training, the LDLR strategy enhances model convergence and sensitivity to early-stage caries by adaptively adjusting the loss weights for challenging and ambiguous lesion samples.


\section{Experiments}
\subsection{Dataset and Evaluation Metrics}
We evaluate the proposed Caries-DETR on two public datasets (AlphaDent~\cite{sosnin2025alphadentdatasetautomatedtooth} and DentalAI~\cite{Dentalai}).
AlphaDent comprises 1,455 intraoral images, which are annotated with pixel-level labels for nine fine-grained categories, including common dental restorations (Abrasion, Filling, Crown) and a detailed classification of caries severity and location (Caries 1 to Caries 6). 
We split the dataset into training (1,237 images), validation (83 images), and testing (135 images) subsets.
DentalAI consists of 2,495 intraoral images collected from diverse clinical scenarios, comprising four categories (Tooth, Caries, Cavity, and Crack).
We split the dataset into training (1,991 images), validation (254 images), and testing (250 images) subsets. 

In addition, we constructed a large-scale intraoral dataset from four publicly available datasets, which is used to train the proposed TSQI.
The details of the dataset are given in the supplementary material (Table~S1).
It comprises 30,943 intraoral photographs, which integrate dental cavities, tooth decay, general intraoral photography, and oral diseases, thereby capturing substantial variability in imaging conditions, tooth morphology, and lesion appearance. 
This heterogeneous corpus allows the model to learn robust domain-specific structural priors before undergoing supervised fine-tuning. 
Crucially, this pre-training dataset was used exclusively for representation learning and has no overlap with the supervised splits of AlphaDent or DentalAI, thus ensuring a fair evaluation without data leakage.

\begin{table*}[!h]
	\centering
	\renewcommand{\arraystretch}{1.2} 
	\resizebox{1\textwidth}{!}{%
		\begin{tabular}{ccc|ccccccccc|ccc}
			\toprule
			\textbf{LIPP} & \textbf{TSQI} & \textbf{LDLR} & 
			\textbf{Abrasion} & \textbf{Filling} & \textbf{Crown} & 
			\textbf{Caries 1} & \textbf{Caries 2} & \textbf{Caries 3} & 
			\textbf{Caries 4} & \textbf{Caries 5} & \textbf{Caries 6} & 
			\textbf{mAP} & \textbf{mAP$_{50}$} & \textbf{mAP$_{75}$} \\
			\midrule
			
			\redcross & \redcross & \redcross & 
			\textbf{\textcolor{blue}{74.2}} & 39.7 & 66.2 & 21.1 & 9.1 & 1.9 & 1.8 & 20.3 & 0.4 & 
			27.2 & 43.2 & 28.2 \\
			
			\greencheck & \redcross & \redcross & 
			71.0 & 37.6 & \textbf{\textcolor{blue}{76.4}} & 17.3 & 9.7 & \textbf{\textcolor{blue}{7.5}} & \textbf{\textcolor{blue}{15.7}} & 17.1 & 1.0 & 
			28.1 & 44.7 & \textbf{\textcolor{red}{29.8}} \\
			
			\redcross & \greencheck & \redcross & 
			\textbf{\textcolor{red}{75.4}} & \textbf{\textcolor{blue}{39.9}} & 74.8 & \textbf{\textcolor{red}{24.8}} & \textbf{\textcolor{red}{10.6}} & 3.0 & 0.3 & \textbf{\textcolor{blue}{20.6}} & \textbf{\textcolor{red}{11.4}} & 
			\textbf{\textcolor{blue}{29.0}} & 46.3 & 26.9 \\
			
			\redcross & \redcross & \greencheck & 
			73.9 & \textbf{\textcolor{red}{40.1}} & 76.2 & \textbf{\textcolor{blue}{23.0}} & 8.8 & 3.8 & 3.0 & 19.8 & 2.1 & 
			27.9 & 44.7 & 28.0 \\
			
			\greencheck & \greencheck & \redcross & 
			72.6 & 39.6 & 72.4 & 20.7 & 9.1 & 3.8 & 6.1 & 19.5 & \textbf{\textcolor{blue}{8.0}} & 
			28.0 & 46.0 & 26.7 \\
			
			\greencheck & \redcross & \greencheck & 
			72.0 & 36.5 & 68.2 & 18.8 & 9.1 & 1.7 & 14.0 & 19.3 & 0.5 & 
			26.7 & 42.3 & 26.1 \\
			
			\redcross & \greencheck & \greencheck & 
			71.4 & 39.4 & 71.6 & 21.8 & \textbf{\textcolor{blue}{9.8}} & \textbf{\textcolor{red}{8.0}} & 14.8 & \textbf{\textcolor{red}{20.7}} & 1.2 & 
			28.7 & \textbf{\textcolor{blue}{48.6}} & 26.5 \\
			
			\greencheck & \greencheck & \greencheck & 
			71.7 & 39.4 & \textbf{\textcolor{red}{76.5}} & 20.2 & 9.4 & 4.7 & \textbf{\textcolor{red}{20.9}} & 18.8 & 7.6 & 
			\textbf{\textcolor{red}{29.9}} & \textbf{\textcolor{red}{50.5}} & \textbf{\textcolor{blue}{29.6}} \\
			\bottomrule
		\end{tabular}%
	}
	\caption{Ablation studies of different modules. LIPP, TSQI and LDLR denote large-scale intraoral photograph pre-training, tooth structure-aware query initialization, and lesion-aware dynamic loss refinement, respectively.
	}
	\label{tab:ablation_studies}
\end{table*}

\begin{table*}[!h]
	\centering
	
	\resizebox{\textwidth}{!}{%
		\begin{tabular}{lc|ccccccccc|c}
			\toprule
			\textbf{Model} & \textbf{Backbone} & 
			\textbf{Abrasion} & \textbf{Filling} & \textbf{Crown} & 
			\textbf{Caries 1} & \textbf{Caries 2} & \textbf{Caries 3} & 
			\textbf{Caries 4} & \textbf{Caries 5} & \textbf{Caries 6} & 
			\textbf{mAP} \\
			\midrule
			\rowcolor[gray]{0.9}\multicolumn{12}{l}{\textit{\textbf{CNN-based Detectors}}} \\
			Faster R-CNN\textit{\textsubscript{NeurIPS'2015}}~\cite{ren2015faster} & R-50          & 65.8 & 30.1 & 60.9 & 12.0 & 8.9 & 2.1 & 0.0 & 18.1 & 2.8 & 22.4  \\
			RetinaNet\textit{\textsubscript{ICCV'2017}}~\cite{lin2017focal}& R-50  & 68.6 & 33.4 & 43.7 & 19.0 & 6.8 & 2.7 & 0.0 & 20.8 & 1.7 & 21.9  \\
			Cascade R-CNN\textit{\textsubscript{TPAMI'2019}}~\cite{cai2019cascade} & R-50         & 62.9 & 32.0 & 61.0 & 12.1 & 5.4 & 2.3 & 0.0 & 14.2 & 2.1 & 21.7  \\
			Grid R-CNN \textit{\textsubscript{CVPR'2019}}~\cite{lu2019grid}& R-50  & 64.4 & 30.8 & 62.3 & 13.7 & 4.9 & 1.4 & 0.0 & 15.0 & 0.9 & 21.5 \\
			FSAF \textit{\textsubscript{CVPR'2019}}~\cite{zhu2019feature}& R-50  & 70.7 & 35.8 & 50.9 & 17.2 & 8.4 & 2.5 & 0.0 & 18.8 & 1.4 & 22.9 \\
			CenterNet \textit{\textsubscript{CVPR'2019}}~\cite{zhou2019objects}& R-50  & 72.0 & 39.1 & 61.4 & 20.7 & 7.8 & 4.1 & 0.0 & 19.5 & 0.0 & 25.0\\
			Dynamic R-CNN \textit{\textsubscript{ECCV'2020}}~\cite{zhang2020dynamic}& R-50  & 70.7 & 35.3 & 64.8 & 17.2 & 8.2 & 1.8 & 0.0 & 16.6 & 0.0 & 23.9\\
			Sparse R-CNN \textit{\textsubscript{CVPR'2021}}~\cite{sun2021sparse}& R-50  & 69.5 & 32.2 & 48.8 & 13.7 & 8.0 & 0.7 & 0.0 & 15.4 & 0.7 & 21.0 \\
			YOLOF \textit{\textsubscript{CVPR'2021}}~\cite{chen2021you}& R-50  & 67.6 & 23.0 & 47.9 & 10.9 & 5.8 & 1.5 & 0.4 & 12.3 & 1.0 & 18.9 \\
			TOOD \textit{\textsubscript{ICCV'2021}}~\cite{feng2021tood}& R-50  & 69.0 & 37.4 & 53.6 & 20.1 & 8.9 & 1.4 & 0.3 & 20.4 & 0.8 & 23.5 \\
			DDOD \textit{\textsubscript{ACM MM'2021}}~\cite{chen2021disentangle}& R-50  & 69.1 & 37.4 & 53.8 & 18.4 & 10.1 & 2.6 & 0.4 & 19.3 & 0.0 & 23.5 \\
			YOLOv12\textit{\textsubscript{{NeurIPS'2025}}}~\cite{tian2025yolov12} & - & 74.6 & 39.6 & 65.8 & 19.7 & \textbf{\textcolor{blue}{12.2}} & 1.8 & 15.3 & 20.4 & \textbf{\textcolor{red}{9.6}} & \textbf{\textcolor{blue}{28.8}}  \\
			\midrule
			\rowcolor[gray]{0.9}\multicolumn{12}{l}{\textit{\textbf{Transformer-based Detectors}}}\\
			DETR\textit{\textsubscript{{ECCV'2020}}}~\cite{carion2020end} & R-50 & 57.5 & 19.8 & 51.9 & 5.1 & 4.7 & 0.7 & 0.0 & 8.3 & 0.0 & 17.0  \\
			Deformable DETR\textit{\textsubscript{{ICLR'2021}}}~\cite{zhu2021deformable} & R-50 & 60.2 & 34.8 & 60.7 & 16.5 & 7.5 & 2.5 &  0.1 & 17.5 & 1.6 & 23.7  \\
			Conditional DETR\textit{\textsubscript{{ICCV'2021}}} ~\cite{meng2021conditional} & R-50 & 68.8 & 29.2 & 66.8 & 18.6 & 9.4 & 3.5 & 1.3 & 14.4 & 1.8 & 23.8 \\
			DAB-DETR\textit{\textsubscript{{ICLR'2022}}} ~\cite{liu2022dab}  & R-50 & 70.6 & 31.8 & 69.9 & 14.3 & 7.3 & 1.4 & 1.4 & 22.1 & 1.2 & 24.5 \\
			DINO\textit{\textsubscript{{ICLR'2023}}}~\cite{zhang2022dino} & R-50 & 74.2 & 39.7 & 66.2 & 21.1 & 9.1 & 1.9 & 1.8 & 20.3 & 0.4 & 27.2  \\
			DDQ DETR\textit{\textsubscript{{CVPR'2023}}}~\cite{zhang2023dense} & R-50 & 73.0 & \textbf{\textcolor{blue}{41.8}} & 63.7 & \textbf{\textcolor{blue}{21.3}} & 10.1 & 4.7 & 0.1 & 20.7 & 1.1 & 26.3  \\
			Co-DETR\textit{\textsubscript{{ICCV'2023}}}~\cite{zong2023detrs} & R-50 & 70.1 & \textbf{\textcolor{red}{42.0}} & 71.9 & 21.0 & 11.5 & \textbf{\textcolor{red}{5.6}} & 0.2 & 20.4 & 0.3 & 27.0 \\
			Grounding DINO\textit{\textsubscript{{ECCV'2024}}}~\cite{liu2024grounding} & R-50 & \textbf{\textcolor{blue}{74.9}} & 41.3 & 65.3 & \textbf{\textcolor{red}{25.2}} & 9.4 & \textbf{\textcolor{blue}{5.5}} & 0.9 & 17.8 & 2.9 & 27.0  \\
			Salience DETR\textit{\textsubscript{{CVPR'2024}}}~\cite{hou2024salience} & R-50 & 69.3 & 25.7 & 60.7 & 10.4 & 6.9 & 3.3 & 12.0 & 12.3 & 4.1 & 22.7 \\
			DEIMv2 \textit{\textsubscript{{arXiv'2025}}}~\cite{huang2025deimv2} & -  & \textbf{\textcolor{red}{75.4}} & 33.8 & \textbf{\textcolor{blue}{74.0}} & 15.5 & 9.3 & 2.6 & \textbf{\textcolor{blue}{17.8}} & 17.4 & 0.6 & 27.4\\ 
			Mr.DETR\textit{\textsubscript{{CVPR'2025}}}~\cite{zhang2024mr} & R-50 & 65.4 & 38.1 & 64.0 & 14.7 & 9.1 & 5.4 & 0.1 & \textbf{\textcolor{blue}{22.5}} & 2.6 & 24.6  \\
			RF-DETR \textit{\textsubscript{{arXiv'2025}}}~\cite{robinson2025rfdetrneuralarchitecturesearch} & R-50 &  \textbf{\textcolor{blue}{74.9}} & 33.7 & 69.4 & 19.1 & \textbf{\textcolor{red}{12.3}} & 3.5 & 16.2 & \textbf{\textcolor{red}{23.3}} & 7.0 & \textbf{\textcolor{blue}{28.8}}  \\
			Caries-DETR(ours)  & R-50 & 71.7 & 39.4 & \textbf{\textcolor{red}{76.5}} & 20.2 & 9.4 & 4.7 & \textbf{\textcolor{red}{20.9}} & 18.8 & \textbf{\textcolor{blue}{7.6}} & \textbf{\textcolor{red}{29.9}} \\
			\bottomrule
		\end{tabular}
	}
	\caption{Quantitative comparison with state-of-the-art detectors on the AlphaDent dataset using the ResNet-50 (R-50) backbone. 
		`-' denotes specialized architectures, \textbf{R-ELAN} for YOLOv12 and \textbf{DINOv3-ViT} for DEIMv2.}
	\label{tab:r-50}
\end{table*}

\begin{table*}[!ht]
	\centering
	
	\resizebox{\textwidth}{!}{%
		\begin{tabular}{lc|ccccccccc|c}
			\toprule
			\textbf{Model} & \textbf{Backbone} & 
			\textbf{Abrasion} & \textbf{Filling} & \textbf{Crown} & 
			\textbf{Caries 1} & \textbf{Caries 2} & \textbf{Caries 3} & 
			\textbf{Caries 4} & \textbf{Caries 5} & \textbf{Caries 6} & 
			\textbf{mAP}  \\
			\midrule
			DINO\textit{\textsubscript{{ICLR'2023}}}~\cite{zhang2022dino} & Swin-L & \textbf{\textcolor{red}{77.0}} & 44.3 & 76.2 & 24.1 & \textbf{\textcolor{red}{12.2}} & 4.4 & \textbf{\textcolor{red}{1.5}} & 23.9 & 3.4 & \textbf{\textcolor{blue}{30.0}}  \\
			DDQ DETR\textit{\textsubscript{{CVPR'2023}}}~\cite{zhang2023dense} & Swin-L & \textbf{\textcolor{blue}{76.6}} & 44.8 & \textbf{\textcolor{red}{76.9}} & \textbf{\textcolor{red}{25.5}} & \textbf{\textcolor{blue}{11.9}} & \textbf{\textcolor{red}{6.1}} & \textbf{\textcolor{blue}{0.7}} & \textbf{\textcolor{blue}{24.3}} & 2.7 & 29.9 \\
			Co-DETR\textit{\textsubscript{{ICCV'2023}}}~\cite{zong2023detrs} & Swin-L & 74.8 & \textbf{\textcolor{red}{47.2}} & \textbf{\textcolor{red}{76.9}} & \textbf{\textcolor{blue}{24.9}} & 11.4 & 4.6 & 0.5 & \textbf{\textcolor{red}{24.8}} & \textbf{\textcolor{blue}{4.3}} & 29.9  \\
			
			Caries-DETR(ours) & Swin-L & 76.1 & \textbf{\textcolor{blue}{45.2}} & \textbf{\textcolor{blue}{76.6}} & 24.5 & 10.4 & \textbf{\textcolor{blue}{5.0}} & 0.5 & 23.7 & \textbf{\textcolor{red}{11.3}} & \textbf{\textcolor{red}{30.4}}\\
			\bottomrule
		\end{tabular}
	}
	\caption{Quantitative comparison with state-of-the-art detectors on the AlphaDent dataset using the Swin-Large (Swin-L) backbone.
	}
	\label{tab:swim-L}
\end{table*}

\paragraph{Evaluation Metrics.}  
For quantitative evaluation, we adopt the standard COCO object detection metrics. Our primary metric is the Mean Average Precision (mAP) averaged over IoU thresholds ranging from 0.50 to 0.95, which reflects overall detection accuracy. To enable a more detailed analysis, we also compute the Average Precision (AP) for each of the nine individual classes in the AlphaDent dataset, which allows us to evaluate the model’s performance on specific dental conditions—especially on the subtle caries classes.
\textcolor{red}{Red} indicates the best result, and \textcolor{blue}{Blue} indicates the second best.

\subsection{Implementation Details}
Our Caries-DETR model is built upon the DINO framework.
Training consists of two phases: 1) the first phase involves the large-scale intraoral photograph pretraining. 2) The second phase involves fine-tuning the proposed Caries-DETR. 
For pretraining, all images are resized to 800×800, the backbone and neck are frozen, and only the encoder is optimized using an L1 loss. We employ the AdamW optimizer with a learning rate of $1 \times 10^{-4}$, training for 50 epochs with a batch size of 16. 
During fine-tuning, the pretrained weights of the first phase and ImageNet are applied.
Training is performed using the AdamW optimizer with a base learning rate of $2 \times 10^{-4}$ and a weight decay of $1 \times 10^{-4}$, for 50 epochs with a total batch size of 8. 
A multi-step learning rate schedule is adopted, decaying the learning rate by a factor of 0.1 at epoch 11. 
The multi-scale resizing (ranging from 480 to 800 pixels on the shorter side with a maximum of 1333 pixels on the longer side) and random horizontal flipping are applied, while at inference, images are resized to a fixed size of 800×800 pixels.

\subsection{Ablation Study}
We conduct an ablation study on the AlphaDent dataset to quantify the contributions of our three novel components. Results are summarized in Table~\ref{tab:ablation_studies}.
LIPP learns robust dental representations from over 30k intraoral images, raising mAP from 27.2\% to 28.1\%. TSQI injects structural priors into query initialization, further improving mAP to 29.0\% and notably boosting performance on challenging caries classes (e.g., Caries 6 AP rises from 0.4\% to 11.4\%), confirming the importance of anatomical guidance. LDLR employs a quality-driven hard mining strategy to address class difficulty imbalance, lifting mAP to 27.9\% and doubling AP on Caries 3 (1.9\% → 3.8\%), showing its effectiveness on hard examples.
Integrating all three components yields the best mAP of 29.9\%, a 2.7\% gain over the baseline, with particularly strong gains on the hardest categories (Caries 4: 20.9\% AP; Caries 6: 7.6\% AP). These results demonstrate the effectiveness of our components. 


\subsection{Quantitative Analysis}
\paragraph{Comparison to the Mainstream Detectors.}
To demonstrate the superior performance of our Caries-DETR, we compared it with the mainstream object detectors on the AlphaDent dataset.
From Table~\ref{tab:r-50} we can observe that Caries-DETR achieves a new state-of-the-art mAP of 29.9\%, significantly outperforming both classic and recent detectors.
Compared to CNN-based methods such as Faster R-CNN and RetinaNet, Caries-DETR obtains a gain of over 7.5 mAP, demonstrating its superior ability to detect subtle dental lesions. 
It also surpasses Transformer-based detectors, including DINO by 2.7 mAP and Co-DETR by 2.9 mAP.
Compared to the recent strong baselines such as YOLOv12 and RF-DETR, Caries-DETR outperforms them by 1.1 mAP. 
Moreover, our method excels on the most challenging categories, achieving 20.9\%AP on Caries 4 and 7.6\%AP on Caries 6, where previous detectors often fail due to low contrast and fine structural details.

\begin{figure*}[!h]
	\centering
	\includegraphics[width=1\linewidth]{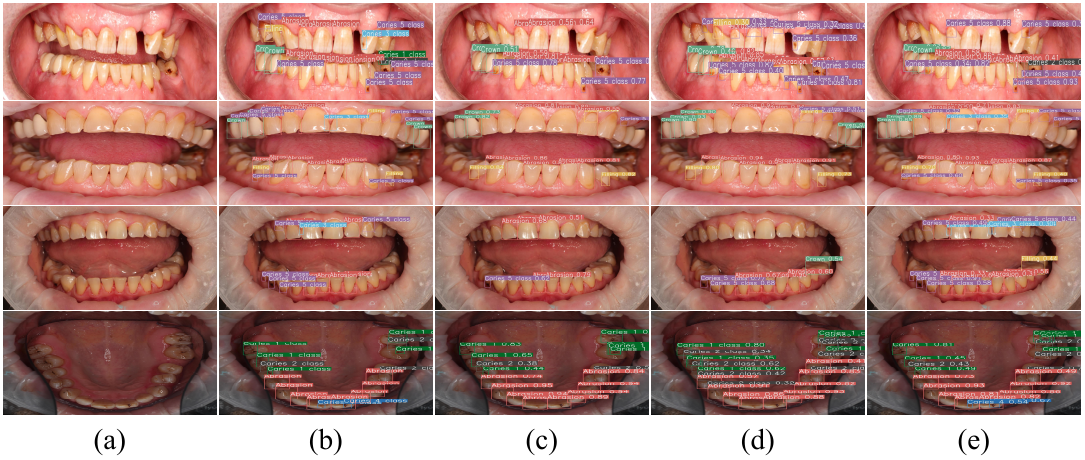}
	\caption{Visual comparison of different methods on the AlphaDent dataset. (a) raw intraoral image (b) Ground truth (c) YOLOv12 (d) RF-DETR (e) Caries-DETR(ours).
	}
	\label{fig:model_column_comparison}
\end{figure*}

\begin{table}[!ht]
	\centering
	
	\resizebox{\columnwidth}{!}{%
		\begin{tabular}{lcccc}
			\toprule
			\textbf{Model} & \textbf{Backbone} & \textbf{mAP} & \textbf{mAP$_{50}$} & \textbf{mAP$_{75}$} \\
			\midrule
			\rowcolor[gray]{0.9}\multicolumn{5}{l}{\textit{\textbf{CNN-based Detectors}}} \\
			CariesFG  \textit{\textsubscript{2023}}~\cite{jiang2023cariesfg} & R-50  & 24.5 & 43.1 & 23.7 \\
			Shafiq et al.\textit{\textsubscript{2023}}~\cite{shafiq2023dental} & R-50  & \textbf{\textcolor{blue}{25.5}} & \textbf{\textcolor{blue}{44.2}} & \textbf{\textcolor{blue}{24.5}} \\
			CariesXplainer\textit{\textsubscript{2025}}~\cite{asghar2025cariesxplainer} & -  & 7.5 & 18.0 & 5.0 \\
			ResFC \textit{\textsubscript{2025}}~\cite{kim2025cnn} & R-50  & 23.9 & 41.9 & 22.4 \\
			M3C \textit{\textsubscript{2025}}~\cite{zhang2025multi} & R-50  & 24.0 & 42.1 & 22.2 \\
			\rowcolor[gray]{0.9}\multicolumn{5}{l}{\textit{\textbf{Transformer-based Detectors}}}\\
			OralTransNet \textit{\textsubscript{2025}}~\cite{asif2025oraltransnet} & - &11.5 & 22.6 & 10.7 \\
			Conformer  \textit{\textsubscript{2026}}~\cite{liu2026enhancing} & R-50  & 22.9 & 39.4 & 22.2  \\
			
			\midrule
			\textbf{Caries-DETR (Ours)} & R-50  &\textbf{\textcolor{red}{29.9}} & \textbf{\textcolor{red}{50.5}} & \textbf{\textcolor{red}{29.6}} \\
			
			\bottomrule
		\end{tabular}%
	}
	\caption{Comparison with caries detection methods on the Alphadent dataset. 
		`-' denotes specialized architectures, \textbf{MobileNetV3} for CariesXplainer and \textbf{MobileNetV2} for OralTransNet.}
	\label{tab:caries_detection_sota_comparison}
\end{table}

\begin{table}[!ht]
	\centering
	
	\resizebox{\columnwidth}{!}{%
		\begin{tabular}{lccccc}
			\toprule
			\textbf{Model} & \textbf{Backbone}  & \textbf{mAP} & \textbf{mAP$_{50}$} & \textbf{mAP$_{75}$} \\
			\midrule
			
			Co-DETR\textit{\textsubscript{ICCV'2023}}~\cite{zong2023detrs} & R-50 & 38.5 & 54.3 & 40.6 \\
			
			DINO\textit{\textsubscript{ICLR'2023}}~\cite{zhang2022dino} & R-50  & 36.2 & 53.4 & 39.6 \\
			
			Grounding DINO\textit{\textsubscript{ECCV'2024}}~\cite{liu2024grounding} & R-50 & 37.0 & 54.4 & 40.0 \\
			
			Salience DETR\textit{\textsubscript{CVPR'2024}}~\cite{hou2024salience} & R-50 & 33.1 & 49.1 & 33.0 \\
			
			DEIMv2\textit{\textsubscript{arXiv'2025}}~\cite{huang2025deimv2} & - & \textbf{\textcolor{blue}{39.1}} & 55.3 & \textbf{\textcolor{red}{41.0}} \\
			
			Mr.DETR\textit{\textsubscript{CVPR'2025}}~\cite{zhang2024mr} & R-50  & 37.2 & 54.3 & 38.6 \\
			
			YOLOv12\textit{\textsubscript{NeurIPS'2025}}~\cite{tian2025yolov12} & - & 38.2 & 53.3 & 40.1 \\
			
			RF-DETR\textit{\textsubscript{arXiv'2025}}~\cite{robinson2025rfdetrneuralarchitecturesearch} & R-50  & 38.7 & \textbf{\textcolor{blue}{56.7}} & 38.5 \\
			
			\midrule
			\textbf{Caries-DETR (Ours)} & R-50 & \textbf{\textcolor{red}{39.2}} & \textbf{\textcolor{red}{58.9}} & \textbf{\textcolor{blue}{40.7}} \\
			
			\bottomrule
		\end{tabular}%
	}
	\caption{Comparison with state-of-the-art methods on the DentalAI dataset. 
		`-' denotes specialized architectures, \textbf{R-ELAN} for YOLOv12 and \textbf{DINOv3-ViT} for DEIMv2.}
	\label{tab:dentalai_sota_comparison}
\end{table}

\paragraph{Stronger Backbone.}
To validate the scalability of Caries-DETR, we evaluate it with a more powerful Transformer backbone, Swin-Large. 
As shown in Table~\ref{tab:swim-L}, while state-of-the-art baselines such as DINO and Co-DETR achieve a high performance plateau of around 30.0\% mAP, Caries-DETR attains the best overall performance with 30.4\% mAP, surpassing DINO by 0.4 mAP and Co-DETR by 0.5 mAP.
More importantly, Caries-DETR achieves 11.3\% AP on the hardest category (Caries 6), which is a relative improvement of more than 7.9 AP over DINO (3.4\% AP). 
This result demonstrates that our Caries-DETR remains effective and complementary to a stronger backbone, achieving good scalability.

\paragraph{Visual Analysis.}
To provide deeper insight into the detection performance of Caries‑DETR, we conducted a visual comparison with two leading methods—YOLOv12 and RF‑DETR on challenging examples in the AlphaDent dataset. 
Visualizations are given in Fig.~\ref{fig:model_column_comparison}.
The selected case is characterized by low contrast and multiple coexisting dental conditions. In columns (c) and (d), YOLOv12 and RF‑DETR fail to detect the Caries 2 lesion (first row, marked in the black box) and the Caries 3 lesion (second and third row, blue box), likely due to the weak boundary features and small scale of early‑stage pathology. Moreover, they completely miss the ambiguous Caries 4 lesion (fourth row, dark-blue box) that exhibits significant texture variation. 
In contrast, Caries-DETR successfully identifies and precisely localizes all these challenging instances. It not only captures the fine‑grained structural anomalies of the Caries 2 and Caries 3 lesions missed by the baselines, but also accurately retrieves the ambiguous Caries 4 lesion. This improved sensitivity stems from the proposed TSQI, which enhances feature representations along tooth boundaries and carious regions, allowing the model to discern subtle pathological changes.

\subsection{Comparison with Caries Detection Methods}
To further verify the effectiveness of our Caries-DETR, we compared it with the caries‑specific detectors on the AlphaDent and DentalAI dataset, in Table~\ref{tab:caries_detection_sota_comparison} and Table~\ref{tab:dentalai_sota_comparison}.

From Table~\ref{tab:caries_detection_sota_comparison} we can observe that Caries-DETR achieves 29.9\% mAP, outperforming the previous best method (Shafiq et al.) by 4.4\% mAP and achieving a 5.1\% gain in mAP$_{75}$.
Several recent lightweight caries detectors, such as CariesXplainer (7.5\% mAP) and OralTransNet (11.5\% mAP), perform substantially worse due to the inherent trade‑off between model efficiency and task complexity. 

In Table~\ref{tab:dentalai_sota_comparison}, Caries-DETR also achieves state‑of‑the‑art performance, in which surpasses transformer‑based detectors, DINO, and Grounding DINO by 3.0 and 2.2 mAP, respectively. 
It also exceeds recent DETR variants like Co‑DETR and RF‑DETR by 0.7 and 0.5 mAP. 
Notably, Caries‑DETR still outperforms DEIMv2 (uses a much larger DINOv3‑ViT backbone) in overall mAP.

\section{Conclusion}
In this work, we introduced Caries-DETR, a novel Transformer framework for caries detection in intraoral images. 
Our approach integrates large-scale intraoral photograph pre-training to distill robust anatomical priors from unlabeled data, and the tooth structure-aware query initialization is designed to guide attention towards subtly structured regions.
A lesion-aware dynamic loss refinement is proposed for quality-driven hard example mining. 
Extensive evaluations conducted on two public datasets demonstrate that Caries-DETR achieves a state-of-the-art performance than existing methods and exhibits good generalization and robustness.

{\small
	\bibliographystyle{ieee_fullname}
	\bibliography{egbib}
}

\end{document}